\documentclass[letterpaper]{article} 
\usepackage{aaai2027}  
\usepackage[hyphens]{url}  
\usepackage{graphicx} 
\usepackage{natbib}  
\usepackage{caption} 
\usepackage{algorithm}
\usepackage{algorithmic}
\usepackage{tabularx}
\usepackage{multirow}
\usepackage{subcaption}

\usepackage{newfloat}
\usepackage{listings}
\DeclareCaptionStyle{ruled}{labelfont=normalfont,labelsep=colon,strut=off} 
\floatstyle{ruled}
\newfloat{listing}{tb}{lst}{}
\floatname{listing}{Listing}

\usepackage{booktabs}

\usepackage{amsmath,amssymb}

\usepackage{xcolor}
\usepackage{tikz}
\usetikzlibrary{positioning,arrows.meta}

\title{\textsc{Hear, See, Spike Once}: Highly Efficient NeuroAI Temporal Neural Networks for Low-Precision Multimodal Audio-Visual Classification and Online Learning}
\author{Shanmuga Venkatachalam, Quinn Jacobson, and John Paul Shen}
\affiliations{ECE, Carnegie Mellon University}

\begin{document}

\maketitle

\begin{abstract}
Spiking neural networks for multimodal audio-visual classification reach high accuracy by adopting deep-learning machinery: multi-stage convolutional or transformer encoders trained end-to-end with surrogate-gradient backpropagation through time. These prior works typically rely on full-precision (32-bit) parameters and do not treat hardware efficiency as a primary design objective. In this paper, we propose NeuTNN-AV, a much simpler hardware-amenable multimodal substrate built on the recently proposed NeuroAI Temporal Neural Network (NeuTNN) and demonstrate that it can match or outperform state-of-the-art SNN baselines with a fraction of their synapse counts. NeuTNN neurons carry multiple active dendritic segments. Each input line spikes at most once (temporal coding) and every weight is learned via a local reward-modulated spike-timing dependent plasticity (STDP) rule.
All weights are 4-bit integers and inference uses only comparisons and additions.
On CREMA-D, it reaches 78.6\% with 2.34M synapses establishing a new state-of-the-art SNN accuracy, against 77.14\% for MISNet-XL (31M parameters).
On UrbanSound8K-AV, NeuTNN reaches 98.82\% with 1.38M synapses, against 96.85\% for SMMT (9.99M parameters) and 98.96\% for MISNet-XL (31M parameters).
Weight memory is 23--106$\times$ smaller than the spiking baselines, and the measured synaptic energy per inference is two to three orders of magnitude lower. Furthermore, the proposed NeuTNN learns CREMA-D in a single pass  with 73.8\% accuracy and demonstrates rapid online continual learning by adapting to unseen actors from a few dozen clips.
\end{abstract}

\section{Introduction}

Audio-visual classification aims to classify an event, object or emotion from a sound and an image or video of it. Spiking neural networks (SNNs) \citep{maass1997networks} coupled with always-on sensors are desirable for this task because their activity is sparse and binary, and they map onto neuromorphic processors \citep{davies2018loihi, merolla2014million, brainchip} that consume very low power and energy.
However, state-of-the-art multimodal SNNs among other artificial neural networks (ANNs) largely achieve competitive accuracy by adopting conventional deep-learning techniques, which compromises their hardware efficiency. EMSNN \citep{liu2022emsnn}, SMMT \citep{guo2024smmt} and MISNet \citep{liu2025misnet} combine convolutional or transformer encoders with cross-modal attention,
are trained offline with surrogate gradients and backpropagation through time \citep{neftci2019surrogate}, utilize full-precision parameters, and cannot learn after deployment without retraining from scratch.

On the other hand, temporal neural networks (TNNs) formulated upon a rigorous space-time algebra encode a value in the timing of a single spike \citep{smith2018spacetime,smith2020temporal}. Spike arrival triggers a synaptic response function in the form of a ramp whose height is capped by its weight, i.e., the ramp-no-leak (RNL) response, and a neuron fires when the summed ramps cross a threshold. Learning is local and utilizes spike-timing dependent plasticity (STDP) \citep{bi1998synaptic,masquelier2007unsupervised}, and the arithmetic is all integer-based, which reduces to counters and comparators on a few bits with no multipliers. This has led to efficient digital microarchitectures \citep{nair2021microarchitecture, nair2022tnn7} that have demonstrated state-of-the-art clustering performance within few tens of $\mu$W power \citep{chaudhari2021unsupervised}.

NeuroAI promotes the convergence of neuroscience and artificial intelligence (AI) research for next era of AI advancements  \citep{zador2023neuroai}. Building on the foundation of TNNs, NeuroAI Temporal Neural Network (NeuTNN) \citep{venkatachalam2026neuroai} widens this convergence to include computer systems, with the aim of developing AI systems that are brain-like in their efficiency as well as capabilities. It is a microarchitecture framework for building neuromorphic sensory processing units that adhere to neocortical structure and functional hierarchy, targeting energy-efficient execution of future AI workloads.




\begin{figure}[htbp]
    \centering
    \begin{subfigure}[b]{0.50\columnwidth}
        \centering
         \includegraphics[width=\textwidth]{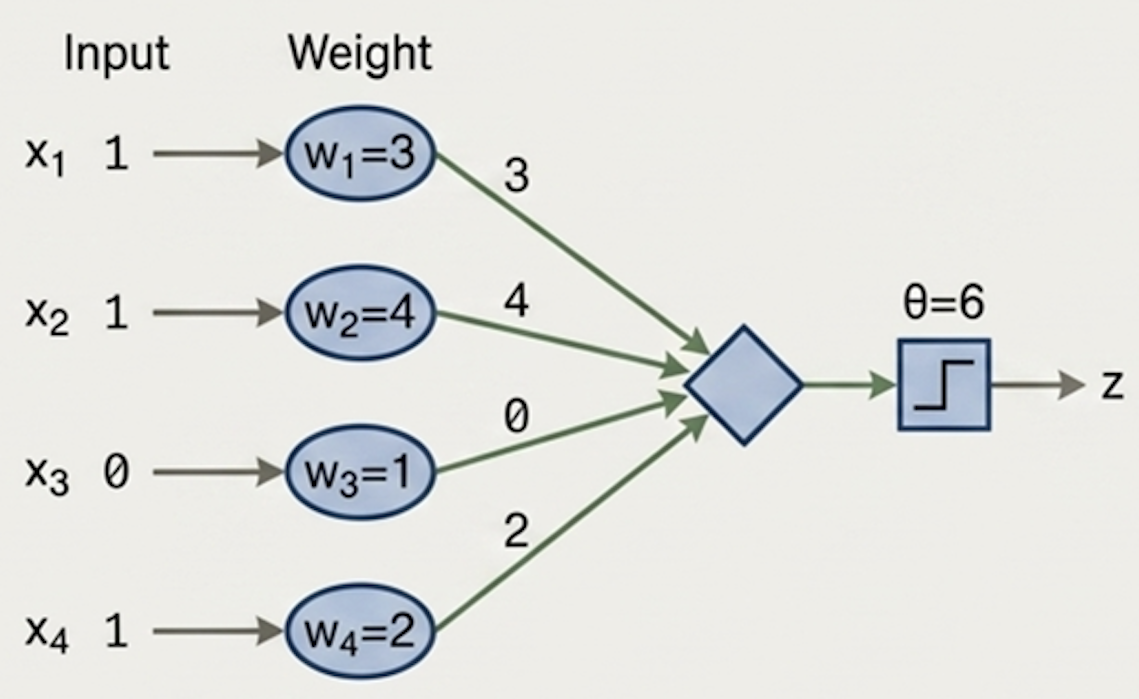}
        \caption{Classic Point Neuron}
        \label{fig:pointneuron}
    \end{subfigure}
    \hfill 
    \begin{subfigure}[b]{0.49\columnwidth}
        \centering
        \includegraphics[width=\textwidth]{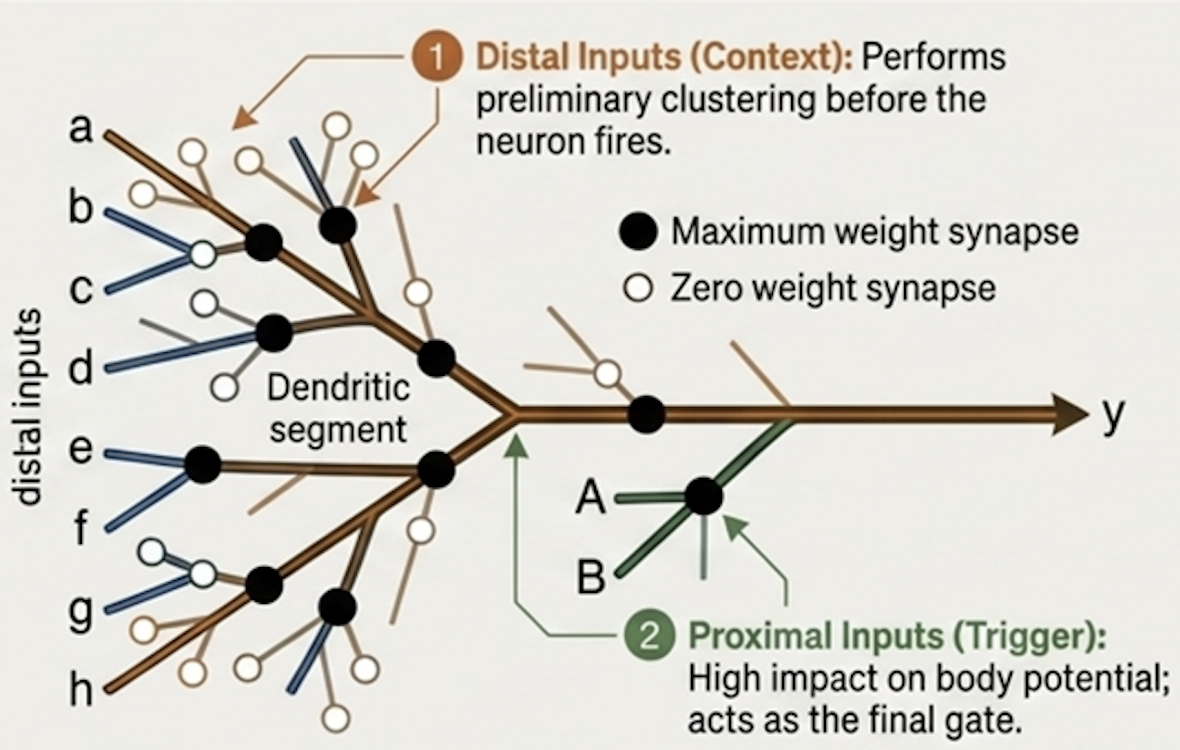}
        \caption{Active Dendrite Model}
        \label{fig:admodel}
    \end{subfigure}
    
    \caption{Point neuron vs. active dendrite neuron.}
    \label{fig:pointvsad}
\end{figure}
Originally, TNN neurons, like most SNNs and ANNs, are point neurons: all synapses feed one sum (Figure \ref{fig:pointneuron}). NeuTNN \citep{venkatachalam2026neuroai} instead uses active-dendrite (AD) neurons (Figure \ref{fig:admodel}). In a biological AD neuron, inputs that arrive near the cell body (proximal) dominate the somatic potential and are required for the neuron to fire. Inputs on the distal segments carry context: each distal segment first clusters its own inputs, and a matching segment prepones the neuron's firing. An AD neuron thus preprocesses and clusters its inputs in several stages before they are accumulated, so a single AD neuron computes like a two-layer network of point neurons \citep{poirazi2003pyramidal,hawkins2016neurons}.
\citet{venkatachalam2026neuroai} show the same benefit in NeuTNN: on MNIST, a single-layer NeuTNN reaches 95\% accuracy with 2.49M synapses, against 94\% for a four-layer TNN with 3.10M.

In this work, we adopt NeuTNN for multimodal audio-visual classification, NeuTNN-AV, and demonstrate that it can match or even outperform deep multimodal SNNs and ANNs on audio-visual benchmarks with a fraction of their synapse counts. In our NeuTNN approach, the sensory inputs arrive on multiple distal segments and the neuron fires with its earliest one, so one neuron can represent several sub-patterns of its class. We use the proximal input as the teaching signal (class label) for supervised learning.
Key contributions are:

\begin{itemize}
  \item \textbf{NeuTNN-AV.} To the best of our knowledge, we propose the first multimodal NeuTNN for audio-visual classification, that fuses the two modalities through a decision layer. Each modality is encoded by corresponding front-ends into spiketimes and fed to corresponding NeuTNN layers, followed by a NeuTNN decision layer. Every stage, including the decision, is an RNL network with integer weights trained by STDP.
  
  \item \textbf{Reward-modulated STDP with recruit-on-error and input-normalized thresholds.} We train each class layer with a reward-modulated STDP rule whose third factor is the class label delivered on each neuron's proximal input. The label sets the sign of every weight update and also decides when the layer adds capacity: a sample's spike pattern is copied into a new distal segment only when the layer misclassifies that sample (\emph{recruit-on-error}). Each segment's firing threshold scales with both the total activity of the input and the segment's total weight.
  
  \item \textbf{Rank-order decision fusion.} Converting each layer's class scores into spike times by value leaves most classes tied. We show that a rank-order code \citep{thorpe1998rank}, in which only each layer's top-ranked classes spike, in rank order, lets a decision layer with a few hundred 4-bit synapses fuse the modalities effectively.

  \item \textbf{Highest reported SNN accuracy on CREMA-D, at a fraction of the cost.} NeuTNN reaches 78.6\% on CREMA-D, 1.5\% above the best reported and the largest SNN (MISNet-XL, 77.1\%), with 13$\times$ fewer synapses. On UrbanSound8K-AV, it comes within 0.14\% of MISNet-XL with 23$\times$ fewer synapses and exceeds all other baselines. All weights use 4 bits instead of 32.
  
  \item \textbf{Online learning and adaptation.} With the same STDP rule, NeuTNN learns CREMA-D in a single pass over the training data and adapts to unseen actors from a few labeled clips, which offline-trained SNNs cannot do without extensive retraining.
\end{itemize}

\section{Background and Related Work}
\label{sec:background}

\paragraph{Temporal neural networks.}
In a TNN, input line $i$ carries one spike at time $t_i\in\{0,\dots,T-1\}$, or none ($t_i=\infty$); earlier spike means stronger input. We write its \emph{earliness} as $a_i=T-t_i$ ($a_i=0$ if silent). A synapse with weight $w\in\{0,\dots,W\}$ responds to its input spike with a ramp that rises by one unit per time step and saturates at $w$. A unit with weight vector $\mathbf{w}$ therefore has potential
\begin{equation}
  V(t)=\sum_i \min\bigl(\max(0,\,t-t_i+1),\,w_i\bigr),
  \label{eq:rnl}
\end{equation}
and fires at the first $t$ with $V(t)\ge\theta$. Here, the weight caps the contribution of its input instead of multiplying it. Winner-take-all (WTA) inhibition lets the earliest unit answer \citep{smith2018spacetime,smith2020temporal}. A NeuTNN \citep{venkatachalam2026neuroai} neuron uses such units as its distal segments (Section~\ref{sec:classlayer}).

\paragraph{STDP rules.}
STDP strengthens synapses whose input precedes the output spike and vice versa \citep{bi1998synaptic}. Unsupervised STDP learns features local to a neuron's synapses \citep{masquelier2007unsupervised,diehl2015unsupervised,kheradpisheh2018stdp} while three-factor rules gate it via an eligibility signal which can be reward or error \citep{izhikevich2007solving,fremaux2016neuromodulated}. Reward-modulated STDP, a specific instance of three-factor rules, has been used for first-spike classification \citep{mozafari2018first}. Orthogonal to STDP, growing the number of prototypes on demand has a long history outside SNNs. Restricted Coulomb Energy (RCE) networks add a prototype when a sample is misclassified \citep{reilly1982neural}, and adaptive resonance theory (ART) adds a category when no existing one matches \citep{carpenter1986adaptive}. Our recruit-on-error rule brings this idea into a spiking dendritic neuron trained using reward-modulated STDP.

\paragraph{Multimodal SNNs and ANNs.}
EMSNN \citep{liu2022emsnn} couples convolutional-recurrent subnetworks through cross-modal attention and targets event-camera and silicon-cochlea recordings, however it does not report on the datasets used
here.
SMMT \citep{guo2024smmt} introduced the UrbanSound8K-AV dataset and fuses convolutional spiking encoders with
spiking cross-attention in the style of Spikformer \citep{zhou2022spikformer}.
MISNet \citep{liu2025misnet} couples the audio and visual streams at every stage through a multimodal leaky-integrate-and-fire (LIF) neuron interleaved with spike-element-wise (SEW)-ResNet stages \citep{fang2021sew} and reports the strongest results on UrbanSound8K-AV and CREMA-D, ahead of ANN methods such as PMR \citep{fan2023pmr} and QMF \citep{zhang2023qmf}.
All of them learn their features end to end via gradient descent.


\section{NeuTNN-AV Architecture}

\begin{figure*}[!t]
\centering
\includegraphics[width=\textwidth]{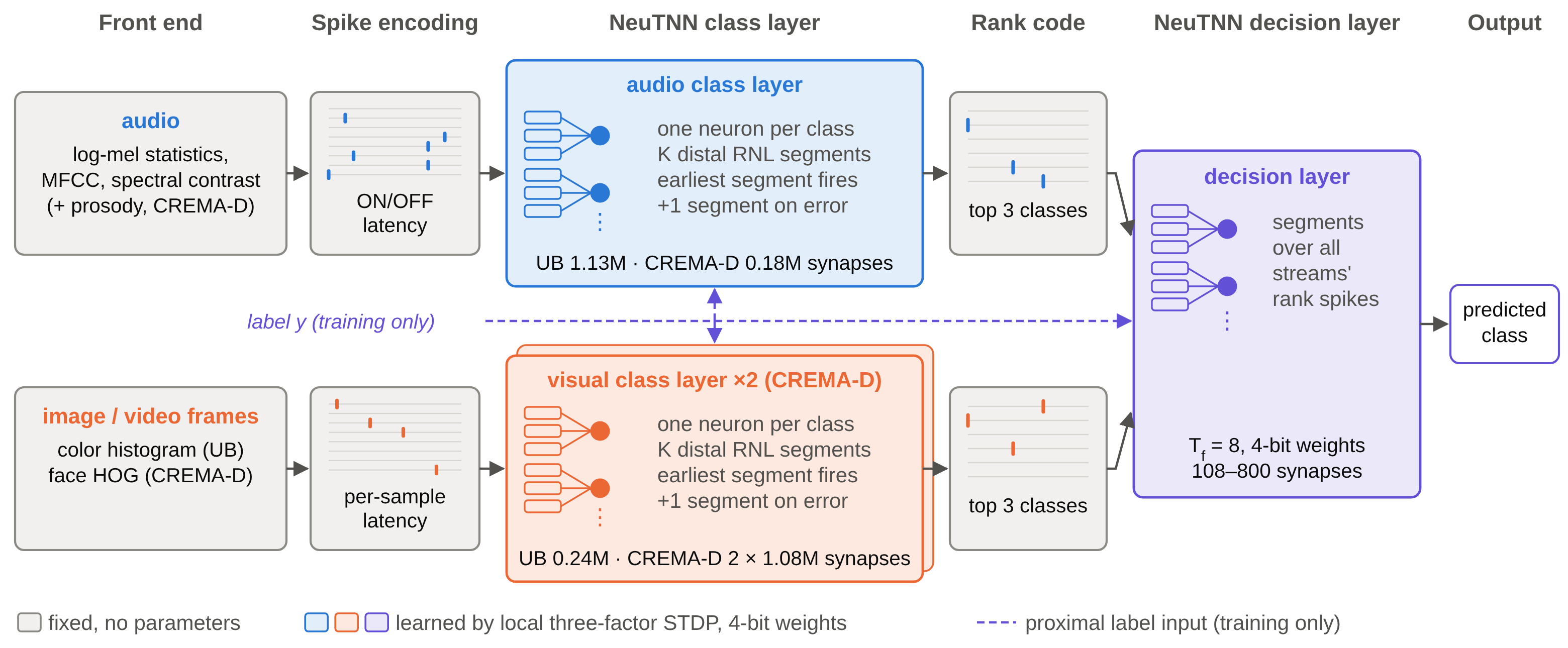}
\caption{NeuTNN-AV: Gray blocks have no learnable parameters: front ends turn each modality into features, and each feature becomes a single spike whose timing encodes its value. Each stream has one NeuTNN class layer (blue: audio; orange: visual) with one neuron per class and up to $K$ distal segments per neuron.
Each class layer passes only its three highest-ranked classes as spikes in rank order to the NeuTNN decision layer (violet), which fuses the streams. The dashed proximal input carries the label during training only. All weights are 4-bit; synapse counts are given per layer for both datasets.}
\label{fig:arch}
\end{figure*}

Figure~\ref{fig:arch} shows the architecture of the proposed NeuTNN-AV. A clip passes through fixed audio and visual front ends, each feature becomes a spiketime, one NeuTNN class layer per stream produces a spike time for every class, and a NeuTNN decision layer reads the rank-order coded class spikes of all streams, thereby fusing the modalities.

\subsection{Front end: Input preprocessing and encoding}
\label{sec:frontend}
The front ends (Table~\ref{tab:features}) are standard signal-processing stages with no learned parameters.

\begin{table}[t]
\centering
\small
\setlength{\tabcolsep}{3pt}
\caption{Audio and visual front end (number of features). 
}
\label{tab:features}
\begin{tabular}{lll}
\toprule
& UrbanSound8K-AV & CREMA-D \\
\midrule
Audio & log-mel statistics (192) & MFCC (240) \\
& MFCC (240) & spectral shape (24) \\
& spectral shape (24) & prosody (16) \\
& & pitch--energy contour (16) \\
& \emph{total 456} & \emph{total 296} \\
\midrule
Vision & HSV color histogram (128) & face HOG, 10 frames (900) \\
\bottomrule
\end{tabular}
\end{table}

\paragraph{Audio.}
The audio descriptors come from a 128-band log-mel spectrogram (22.05\,kHz, 23\,ms hops). Log-mel statistics are the mean, standard deviation, 10th/50th/90th percentiles and maximum of 32 pooled bands.
MFCC features are the mean and standard deviation of 40 coefficients and their first and second differences \citep{davis1980mfcc}; spectral shape is seven-band contrast \citep{jiang2002contrast} plus centroid, bandwidth, roll-off, flatness and zero-crossing rate. For emotion, we add prosody, summary statistics of the pitch \citep{decheveigne2002yin} and energy, and an eight-point contour of log pitch and log energy over the voiced part of the utterance. Voices differ more across actors than across emotions, so CREMA-D audio features are standardised per actor over all of that actor's clips, without labels.

\paragraph{Vision.}
The UrbanSound8K-AV image becomes an $8\times4\times4$ hue-saturation-value (HSV) color histogram.
For CREMA-D, we crop the face in ten frames per clip with a Haar cascade \citep{viola2001rapid},
resize it to $48\times48$ and average the frames' HOG descriptors \citep{dalal2005hog}.

\paragraph{Spike encoding.}
Larger values spike earlier. Audio features are standardised with training-set statistics and can be negative, so each feature $z$ drives two lines, ON for $z\ge0$ and OFF for $z<0$. The active line spikes at
\begin{equation}
  t=\mathrm{round}\Bigl(\bigl(1-\min(|z|,3)/3\bigr)(T-1)\Bigr)
  \label{eq:onoff}
\end{equation}
and the other line is silent, so an audio layer has twice as many lines as features.
For roughly Gaussian features, |z| $\leq$ 3 holds about 99.7\% of values.
Visual features are non-negative and use a per-sample temporal code, $a_i=\mathrm{round}(T\,x_i/\max_j x_j)$: the sample's largest feature spikes first, and scaling the whole vector does not change the code.
We use $T=16$ throughout.

\subsection{NeuTNN class layer}
\label{sec:classlayer}

\begin{figure}[!t]
\centering
\includegraphics[width=\columnwidth]{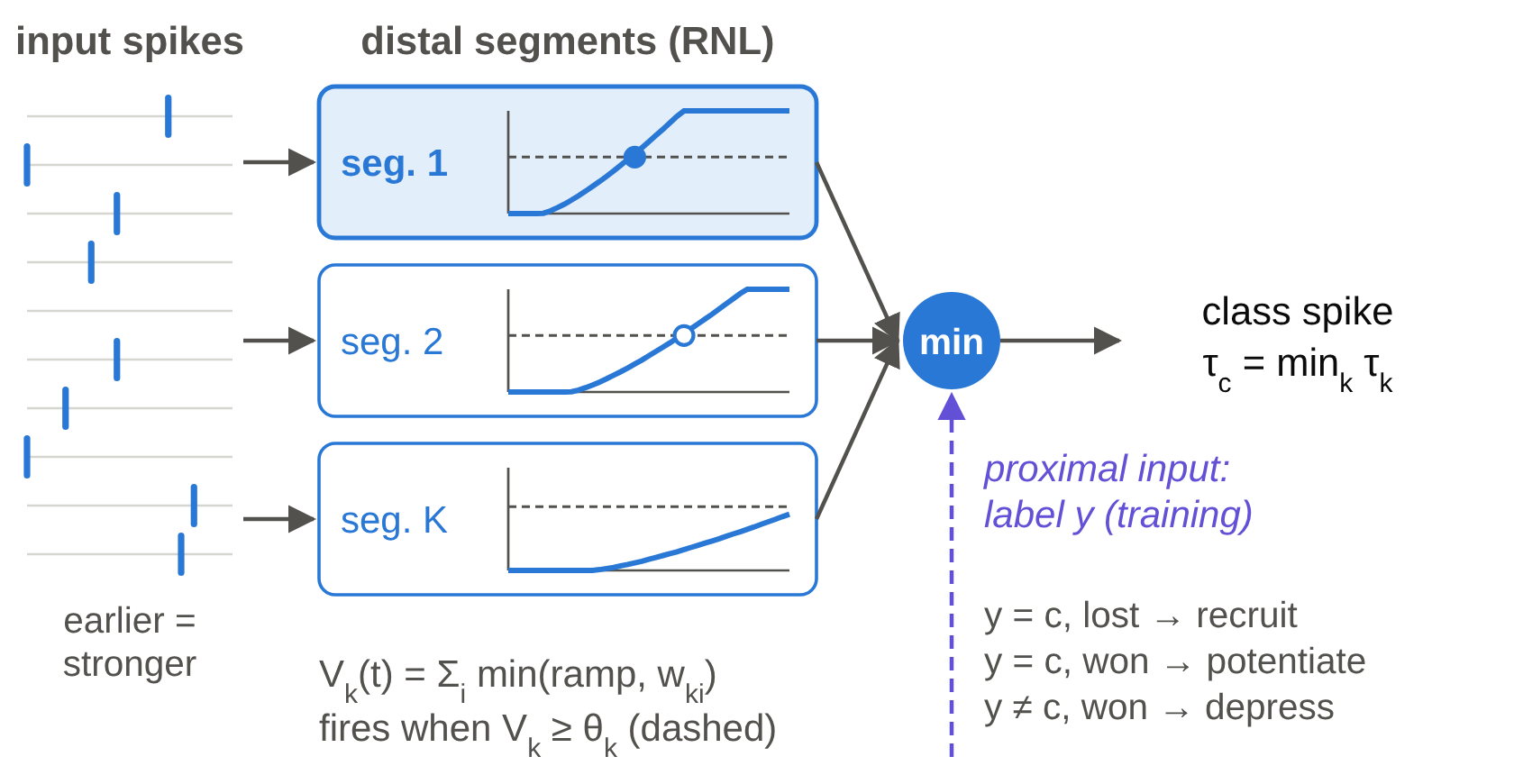}
\caption{One NeuTNN class neuron: Each sensory line spikes once. Every distal segment is a ramp-no-leak unit whose potential rises after its inputs spike; it fires when it reaches its input-normalized threshold $\theta_k$ (dashed). The neuron's class spike is the earliest spiking segment. Proximal input carries the label $y$ during training.
}
\label{fig:neuron}
\end{figure}

A class layer has one neuron per class (Figure~\ref{fig:neuron}). Each neuron has up to $K$ distal segments, all
connected to $P$ sensory input lines, and one proximal input that carries the neuron's class label during training. Segment $k$ computes Eq.~\eqref{eq:rnl} with weights $\mathbf{w}_k$ and fires at $\tau_k$, the
first step at which $V_k(t)\ge\theta_k(\mathbf{a})$. Here, $\mathbf{a}$ is the sample's earliness vector (Section~\ref{sec:background}). Each segment's threshold is recomputed for every sample from two sums: the input's total earliness $\lVert\mathbf{a}\rVert_1=\sum_i a_i$, which is large when many lines spike early, and the segment's total weight $\lVert\mathbf{w}_k\rVert_1=\sum_i w_{ki}$:
\begin{equation}
  \theta_k(\mathbf{a}) = \rho\, max\bigl(\lVert\mathbf{w}_k\rVert_1,\lVert\mathbf{a}\rVert_1\bigr),
  \label{eq:thr}
\end{equation}
where hyperparameter $\rho$ is the fraction of the sum a segment must reach. $\rho$ is one value per layer (0.4--0.85 in our models).

This normalization makes the segments race and compare patterns rather than magnitudes. By step $T-1$, the potential of segment $k$ is the overlap $\sum_i\min(a_i,w_{ki})$ (Eq.~\eqref{eq:rnl} with $a_i=T-t_i$), which is at most the smaller of the two sums and reaches it only when the weights match the input line by line. The segment therefore fires by step $T-1$ if and only if
\begin{equation}
  \frac{\sum_i\min(a_i,w_{ki})}{max\bigl(\lVert\mathbf{w}_k\rVert_1,\lVert\mathbf{a}\rVert_1\bigr)}\ge\rho,
  \label{eq:sim}
\end{equation}
where the left side is a similarity between 0 and 1, in the spirit of a cosine. A fixed threshold would instead let a strong input fire every segment early, and let a segment with large weights fire on almost any input. This acts as divisive normalization \citep{carandini2012normalization} shared by all segments and is computed once per segment, not per synapse.

A neuron fires with its earliest segment, and the earliest neuron gives the class. Segments that fire at the same step are ordered by $V_k/\theta_k$, how far each has passed its own threshold, which picks the segment that crossed first.


\subsection{Reward-modulated STDP with recruit-on-error}
\label{sec:stdp}

Each class neuron has a fixed pool of $K$ physical distal segments. A segment is either idle, contributing nothing and drawing no synaptic work, or active. Training starts with one active segment per neuron, copied from a random training sample of its class, and recruits idle segments from the neuron's pool when the layer errs, so that a class with several distinct modes, such as different speakers or recording conditions, can hold one pattern/prototype per mode. For every training sample the reward-modulated STDP rule decides two things from local signals only: how to change the weights of active segments, and whether to recruit an idle one.

For each training sample with label $y$, the proximal input of neuron $y$ is active and all other proximal inputs are silent. This proximal input is the third factor of the STDP rule: together with the outcome of the distal race, it sets the sign of the weight update (the reward) and decides whether an idle segment is recruited. Let $k^*$ be the winning segment, the first to fire among all neurons' segments, and $k^+$ the earliest segment of neuron $y$. Neuron $y$ \emph{loses} if $k^*$ belongs to another neuron or none of its segments fires. The weight update term for synapse $i$ of segment $k$ is given by $e_{ki}=a_i-w_{ki}$. An input that spiked early pulls its weight toward its earliness, and a silent input decays its weight toward zero. Each sample triggers exactly one of three updates (Algorithm~\ref{alg:learn}):
\begin{itemize}
  \item \emph{Recruit} (proximal active, neuron lost): if neuron $y$ corresponds to the correct label but it lost, has an idle segment and the sample has not been imprinted before, the sample is copied into an idle segment, $\mathbf{w}_{\text{new}}\leftarrow\mathbf{a}$, and nothing else changes. If neuron $y$ has no idle segment, or the sample was imprinted before, it is potentiated as below.
  \item \emph{Potentiate} (proximal active, neuron won): if neuron $y$ corresponds to the correct label and won, reward is $+1$. $\mathbf{w}_{k^*}\leftarrow\mathbf{w}_{k^*}+\eta\,\mathbf{e}_{k^*}$.
  \item \emph{Depress} (proximal silent, neuron won): if neuron $y$ won but does not correspond to the correct label, reward is $-1$. $\mathbf{w}_{k^*i}\leftarrow \mathbf{w}_{k^*i}-\eta\,|\mathbf{e}_{k^*i}|$ for $\mathbf{a}_i>0$.
\end{itemize}

All other segments are unchanged. Weights are rounded to integers in $\{0,\dots,15\}$ (4 bits). 
Note that the rule needs only the current input, the proximal signal and the identity of the winner, and the supervision is local to each neuron rather than an error propagated through layers.

\begin{algorithm}[t]
\caption{Reward-modulated STDP with recruit-on-error, for one training sample (earliness $\mathbf{a}$, label $y$)}
\label{alg:learn}
\begin{algorithmic}[1]
\STATE run the distal race; $k^*\leftarrow$ winning segment, $k^+\leftarrow$ earliest segment of neuron $y$
\IF{($\mathrm{class}(k^*)\neq y$ \OR no segment of $y$ fired) \AND $y$ has a free segment \AND $\mathbf{a}$ not yet
imprinted}
  \STATE $\mathbf{w}_{\text{free}}\leftarrow\mathbf{a}$ \COMMENT{recruit a distal segment}
\ELSIF{$\mathrm{class}(k^*)=y$}
  \STATE $\mathbf{w}_{k^*}\leftarrow\mathbf{w}_{k^*}+\eta(\mathbf{a}-\mathbf{w}_{k^*})$ \COMMENT{reward $+1$}
\ELSE
  \STATE $w_{k^*i}\leftarrow w_{k^*i}-\eta|a_i-w_{k^*i}|$ for all $i$ with $a_i>0$ \COMMENT{reward $-1$}
  \STATE $\mathbf{w}_{k^+}\leftarrow\mathbf{w}_{k^+}+\eta(\mathbf{a}-\mathbf{w}_{k^+})$
\ENDIF
\STATE clip weights to $[0,T-1]$, round and cap at 15
\end{algorithmic}
\end{algorithm}

\subsection{Rank-order decision layer}

Each class layer ranks the classes by how early their best segment fires, earliest first, with ties broken by the margin $V/\theta$. The decision layer receives these rankings as spikes in a rank-order code \citep{thorpe1998rank}. In each stream, top $m$ classes are selected as follows: the class ranked $r$ ($r=0$ for the top class) spikes at $\mathrm{round}(T_f\,r/m)$ if $r<m$ and stays silent otherwise. We use $T_f=8$ and $m=3$, so each stream sends at most three spikes.

The rank code suits fusion for several reasons. It needs no calibration between streams, because ranks do not depend on the scale of each layer's scores. It keeps the runner-up classes, which carry much of the complementary evidence: a class that ranks high in both streams can beat one that is first in only one. And it is sparse and short, so the decision layer needs only a few spikes, eight time steps and 4-bit weights. Coding the scores' values directly as latencies instead (i.e., temporal code) gives most classes the same spike time and costs 4.6--12.6\% in accuracy (Table~\ref{tab:ablation}).

The decision layer is a NeuTNN layer with one neuron per class. Each neuron has $K_f$ distal segments over the $L\cdot C$ rank-coded lines of $L$ streams and $C$ classes and a proximal label input, and is trained by the same STDP rule as earlier.
\section{Experimental Setup}

\paragraph{Datasets.}
\emph{UrbanSound8K-AV} \citep{guo2024smmt} pairs 8{,}732 clips of UrbanSound8K \citep{salamon2014urbansound} (10 urban sound classes, up to 4\,s) with images of the same class; we use 70\% for training and 30\% for testing, as SMMT. \emph{CREMA-D} \citep{cao2014cremad} contains 7{,}442 video clips of 91 actors speaking 12 sentences in six emotions (anger, disgust, fear, happiness, neutral, sadness); we use the 6{,}698/744 train/test partition of MISNet. CREMA-D is the harder task: the class is a subtle change of expression and prosody on the same face and sentence, the labels are the actors' intended emotions, which human raters recover in only about 64\% of clips from audio and video together \citep{cao2014cremad}.

\paragraph{Protocol.}
All hyperparameters are chosen on a validation split of 30\% of the training data. The class layers are trained on the remaining 70\% to produce validation scores for the decision layer, whose settings ($K_f\in\{1,2,4\}$, $\rho_f$, $m$) are chosen by two-fold cross-validation inside the validation split; all layers are then retrained on the full training set and evaluated once on the test set. Class layers use $\eta=0.2$
and the ratio $\rho$ is chosen per layer (Table~\ref{tab:config}).

\begin{table}[t]
\centering\small
\setlength{\tabcolsep}{2pt}
\caption{Layer configurations and sizes. Lines: spike input lines per
segment; Seg.: distal segments; Syn.: synapses.}
\label{tab:config}
\begin{tabularx}{\columnwidth}{XXrccrr}
\toprule
Layer & Input & Lines & Code & $\rho$ & Seg. & Syn. \\
\midrule
\multicolumn{7}{l}{\emph{UrbanSound8K-AV}} \\\\
audio & spectral & 912 & ON/OFF & 0.6 & 1{,}237 & 1.13M \\
image & color & 128 & temporal & 0.8 & 1{,}873 & 0.24M \\
decision & 2 streams & 20 & rank & 0.5 & 20 & 400 \\
\midrule
\multicolumn{7}{l}{\emph{CREMA-D}} \\\\
audio & spectral, prosody & 592 & ON/OFF & 0.4 & 300 & 0.18M \\
visual layer 1 & face HOG & 900 & temporal & 0.85 & 1{,}200 & 1.08M \\
visual layer 2 & face HOG & 900 & temporal & 0.8 & 1{,}200 & 1.08M \\
decision & 3 streams & 18 & rank & 0.5 & 6 & 108 \\
\bottomrule
\end{tabularx}
\end{table}

\paragraph{Baselines.}
We compare with the published results of SMMT \citep{guo2024smmt} and MISNet
\citep{liu2025misnet}, and with the ANN methods reported by MISNet. Parameter counts and energies are taken from those
papers. Like them, we exclude the cost of spectrogram and input preprocessing.

\section{Experimental Results}

\begin{table}[t]
\centering\small
\setlength{\tabcolsep}{3.5pt}
\caption{UrbanSound8K-AV accuracy (\%): Baseline parameters are 32-bit. NeuTNN audio and visual columns are its single-modality class layers.}
\label{tab:ub}
\begin{tabular}{lcccc}
\toprule
Method & Audio & Visual & Fused & Params \\
\midrule
SMMT, $T{=}1$ & 92.10 & 90.74 & 94.30 & 9.99M \\
SMMT, $T{=}4$ & -- & -- & 96.85 & 9.99M \\
MISNet-XS & -- & -- & 97.71 & 4.06M \\
MISNet-S & -- & -- & 98.09 & 7.33M \\
MISNet-L & -- & -- & 98.52 & 19.63M \\
MISNet-XL & -- & -- & 98.96 & 31.00M \\
\midrule
\textbf{NeuTNN-AV} & 92.44 & 92.52 & \textbf{98.82} & 1.38M \\
\bottomrule
\end{tabular}
\end{table}

\begin{table}[t]
\centering\small
\setlength{\tabcolsep}{3.5pt}
\caption{CREMA-D accuracy (\%): Baseline numbers from \citet{liu2025misnet} with 32-bit parameters. NeuTNN audio and visual columns are its single-modality class layers (note that there are two visual layers).}
\label{tab:cremad}
\begin{tabular}{lccccc}
\toprule
Method & Type & Audio & Visual & Fused & Params \\
\midrule
PMR & ANN & -- & -- & 61.10 & 22.36M \\
QMF & ANN & -- & -- & 63.71 & -- \\
SMMT & SNN & -- & -- & 62.50 & 9.99M \\
MISNet-XS & SNN & -- & -- & 66.45 & 4.06M \\
MISNet-S & SNN & -- & -- & 68.72 & 7.33M \\
MISNet-L & SNN & -- & -- & 75.22 & 19.63M \\
MISNet-XL & SNN & -- & -- & 77.14 & 31.00M \\
\midrule
\textbf{NeuTNN-AV} & TNN & 52.42 & 71.24 / 65.86 & \textbf{78.63} & 2.34M \\
\bottomrule
\end{tabular}
\end{table}

\paragraph{Accuracy and size comparison.}
Tables~\ref{tab:ub} and~\ref{tab:cremad} summarise the accuracy comparison. On UrbanSound8K-AV, the NeuTNN class layers alone are on par with SMMT's single-modality encoders (92.44\% audio, 92.52\% visual). The spiking decision layer lifts the pair to 98.82\%, nearly matching the largest baseline, MISNet-XL, with a 23$\times$ lower synapse count (3x lower than even the smallest baseline, MISNet-XS) and outperforming all other baselines. On CREMA-D, NeuTNN reaches 78.63\%, 12\% above MISNet-XS with just above half of its size and 1.5\% above MISNet-XL with 13$\times$ fewer synapses. Combining the modalities adds 6.3\% on UrbanSound8K-AV and 7.4\% on CREMA-D over the best unimodal stream, thereby illustrating the efficacy of multimodal fusion.

\begin{figure*}[!t]
\centering
\includegraphics[width=1.03\textwidth]{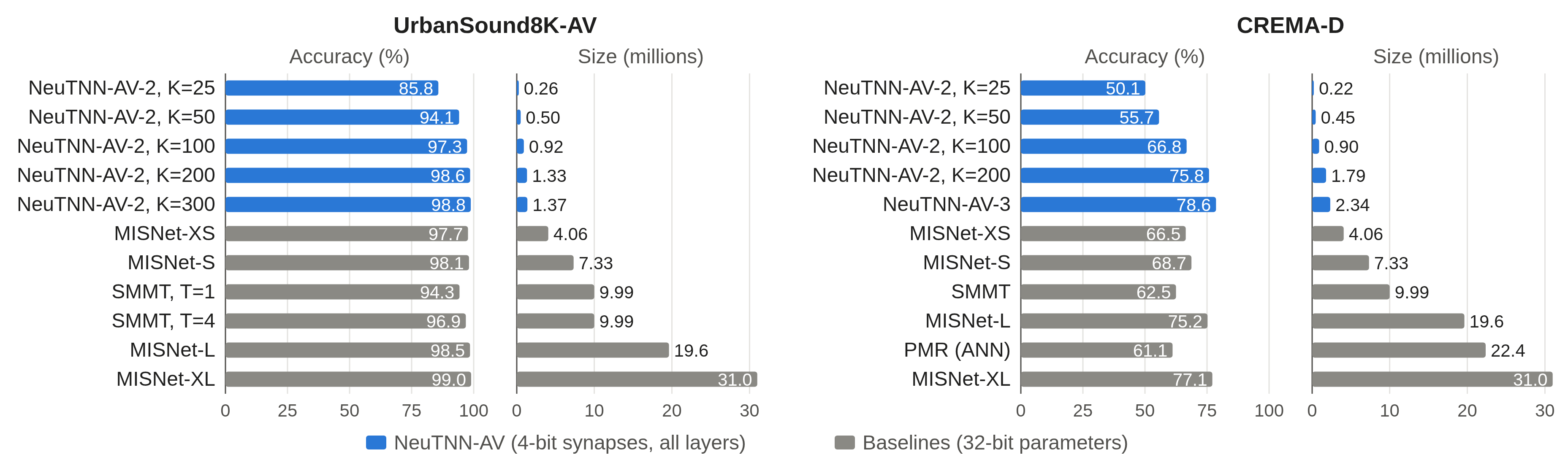}
\caption{Accuracy and model size on UrbanSound8K-AV (left) and CREMA-D (right). For each dataset, the two panels share one row per model, and models are ordered by size. NeuTNN-AV implements 4-bit synapses in all layers, including the decision layer; baselines use 32-bit parameters. NeuTNN-AV-2 has one audio and one visual class layer, each with a pool of $K$ segments per neuron; NeuTNN-AV-3, the final CREMA-D model, adds a second visual layer.}
\label{fig:size}
\end{figure*}

Figure~\ref{fig:size} places the models on a common size axis. On both datasets, NeuTNN demonstrates significantly better accuracy-to-size tradeoff. On UrbanSound8K-AV, it reaches 94\% with under 1M synapses and 98.82\% with 1.38M. On CREMA-D, a two-layer model with 1.79M synapses already matches MISNet-L at 19.6M.

Figure~\ref{fig:size} shows accuracy and size (synapse count in millions) side by side for every model. Every NeuTNN model is smaller than the smallest baseline, and on both datasets, NeuTNN matches the accuracy of baselines that are an order of magnitude larger. On UrbanSound8K-AV, it reaches 94.1\% with 0.50M synapses, 97.3\% with 0.92M, above SMMT at four time steps with 11$\times$ fewer synapses, and 98.8\% with 1.38M, above MISNet-L and 0.2\% below MISNet-XL with 23$\times$ fewer synapses.
On CREMA-D, two-layer NeuTNN with 1.79M synapses (75.8\%) matches MISNet-L at 19.6M, and three-layer NeuTNN with 2.34M (78.6\%) exceeds MISNet-XL at 31M.

\paragraph{Synaptic Distribution.}
In UrbanSound8K-AV, the audio layer recruits about 1{,}250 segments over 912 lines (1.14M synapses) and the image layer about 1{,}860 segments over 128 lines (0.24M). The decision layer has 108--400 synapses. On CREMA-D, the audio layer uses 300 segments over 592 lines (0.18M) and each visual layer 1{,}200 segments over 900 lines (1.08M); the decision layer has 108 synapses. It can be seen that the decision layer is negligible relative to the class layers.

\begin{table}[t]
\centering\small
\setlength{\tabcolsep}{3pt}
\caption{CREMA-D per-class recall (\%) of the audio layer, the best visual layer and the fused 4-bit network, and
the fused confusion counts (rows: true class).}
\label{tab:perclass}
\begin{tabular}{lccc|cccccc}
\toprule
 & aud. & vis. & fused & ANG & DIS & FEA & HAP & NEU & SAD \\
\midrule
ANG & \textbf{76.2} & 61.5 & 80.3 & 98 & 6 & 7 & 4 & 5 & 2 \\
DIS & 33.6 & \textbf{82.4} & 84.8 & 3 & 106 & 5 & 4 & 0 & 7 \\
FEA & 50.8 & 60.8 & 63.1 & 16 & 5 & 82 & 3 & 6 & 18 \\
HAP & 37.8 & \textbf{89.9} & 90.8 & 7 & 1 & 0 & 108 & 2 & 1 \\
NEU & 57.1 & 75.9 & 85.7 & 4 & 0 & 5 & 0 & 96 & 7 \\
SAD & 58.8 & 59.6 & 69.9 & 2 & 7 & 11 & 2 & 19 & 95 \\
\bottomrule
\end{tabular}
\end{table}

\paragraph{Modality Contribution.}
Table~\ref{tab:perclass} breaks CREMA-D down by emotion. The modalities
are complementary in the way one would expect from the signals: audio is the better stream for anger, which is carried
by loudness and pitch, while the face is far better for happiness and disgust, which are carried by the mouth and nose
region. The decision layer improves on the better stream for all six emotions; the largest gains are for neutral
(+10\%) and sadness (+10), where neither stream is reliable alone. Fear and
sadness remain the weakest classes and are mostly confused with each other and with neutral.

\begin{table}[t]
\centering\small
\setlength{\tabcolsep}{2.5pt}
\caption{Model size, weight memory and synaptic energy per inference. Baseline energies are from
\citet{liu2025misnet}. NeuTNN energy is estimated similar to the baselines: measured RNL unit increments per test inference, each charged as one 32-bit
addition (0.9\,pJ, 45\,nm; \citealp{horowitz2014energy}).}
\label{tab:eff}
\begin{tabular}{llcccc}
\toprule
Data & Model & Params & Bits & Memory & Energy \\
\midrule
\multirow{4}{*}{UB8K-AV} & SMMT & 9.99M & 32 & 40.0\,MB & 3.99\,mJ \\
 & MISNet-XS & 4.06M & 32 & 16.2\,MB & 0.97\,mJ \\
 & MISNet-XL & 31.00M & 32 & 124.0\,MB & 4.23\,mJ \\
 & \textbf{NeuTNN} & \textbf{1.38M} & \textbf{4} & \textbf{0.69\,MB} & \textbf{0.0011\,mJ} \\
\midrule
\multirow{4}{*}{CREMA-D} & SMMT & 9.99M & 32 & 40.0\,MB & 5.78\,mJ \\
 & MISNet-S & 7.33M & 32 & 29.3\,MB & 2.41\,mJ \\
 & MISNet-XL & 31.00M & 32 & 124.0\,MB & 6.88\,mJ \\
 & \textbf{NeuTNN} & \textbf{2.34M} & \textbf{4} & \textbf{1.17\,MB} & \textbf{0.0088\,mJ} \\
\bottomrule
\end{tabular}
\end{table}

\paragraph{Hardware Efficiency.}
Table~\ref{tab:eff} compares the hardware cost. Weight memory is 58$\times$ smaller
than SMMT and 23$\times$ smaller than MISNet-XS on UrbanSound8K-AV, and 106$\times$ smaller than MISNet-XL and
25$\times$ smaller than MISNet-S on CREMA-D.

For energy estimation, we adopt a methodology analogous to the one used in \citep{liu2025misnet}.
A synapse whose input has earliness $a_i$ and weight $w_{ki}$ adds $\min(a_i,w_{ki})$ unit steps to its segment's potential, so the total number of unit increments per test clip, summed over all segments, is the synaptic work. On UrbanSound8K-AV it is 1.22M per clip, 0.89 per synapse: half of the audio lines are silent by construction of the ON/OFF code, 68\% of the color-histogram lines are silent, and 43--66\% of the learned weights are zero. On CREMA-D it is 9.7M per clip, 4.2 per synapse, dominated by the face layers, where 94\% of the HOG lines spike.
Charged as 32-bit floating-point accumulates (0.9 pJ), the same cost the baselines are charged, this is 1.1\,$\mu$J and 8.8\,$\mu$J per inference, 880--3{,}850$\times$ below the spiking baselines on UrbanSound8K-AV and 270--780$\times$ on CREMA-D; even if every synapse ramped to its full weight, the totals would be 19 and 32\,$\mu$J.
The comparison is conservative in a second way: an RNL synapse is a 4-bit counter, and an 8-bit addition costs only about 0.03\,pJ rather than 0.9\,pJ \citep{horowitz2014energy}. The remaining operations are one threshold comparison per segment and time step and one multiplication per segment for Eq.~\eqref{eq:thr}, about $10^5$ in total. There is no multiplication per synapse and each input line spikes once, which suits counter-and-comparator implementations of TNNs \citep{nair2021microarchitecture}.


\begin{figure*}[!t]
\centering
\includegraphics[width=0.9\textwidth]{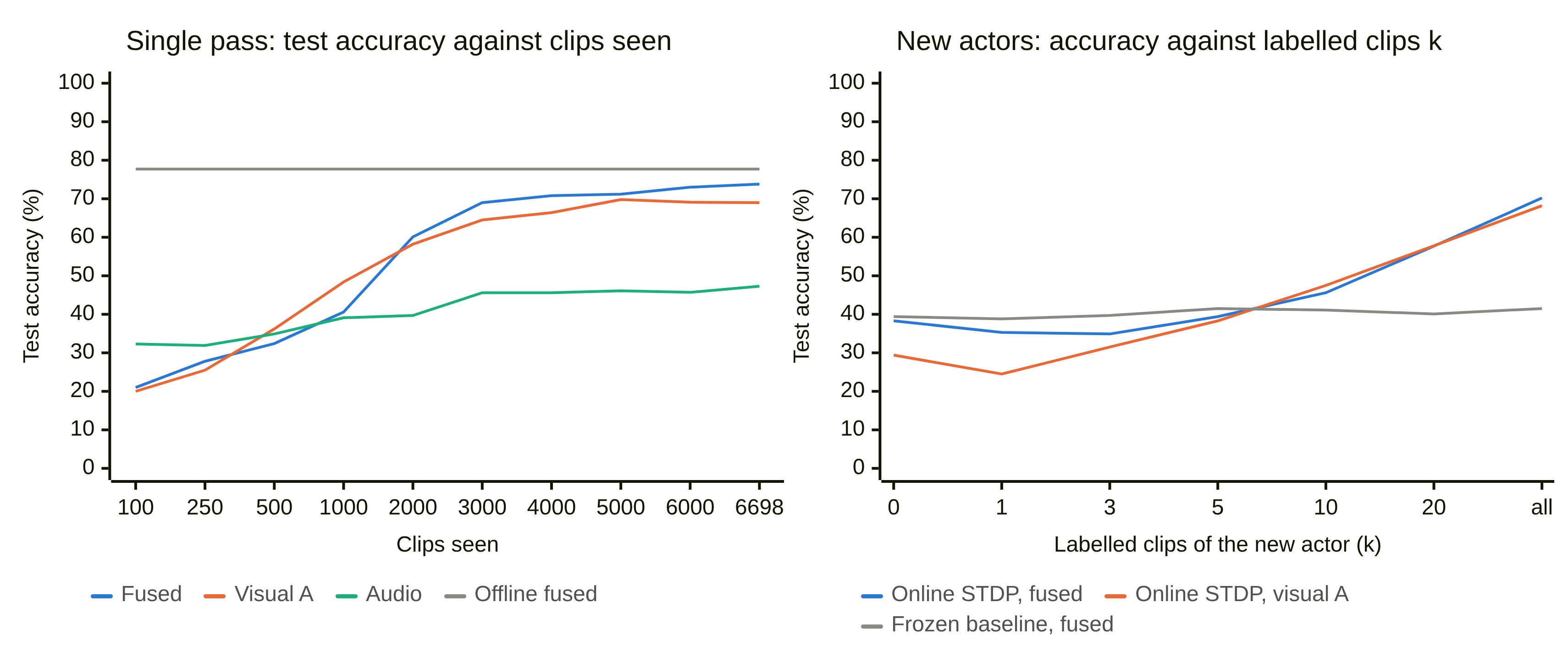}
\caption{Online STDP learning on CREMA-D with single-pass, one-clip-at-a-time STDP. Left: test accuracy against clips seen; the horizontal line is the same configuration trained offline. Right: accuracy on 31 unseen actors after learning from $k$ of their labeled clips (mean over actors), against the same network frozen with only its label-free normalization updated.}
\label{fig:online}
\end{figure*}

\section{Online Learning on CREMA-D}

Because every update in NeuTNN is local and uses one sample, it can learn online. We ran inputs through the CREMA-D model one clip at a time in a single pass, with no epochs and no stored data: each clip is first classified, then learned by the class layers and the decision layer. Statistics that the offline pipeline estimates from the training set are computed causally: per-actor normalization uses the actor's clips seen so far, shrunk toward global statistics.

\paragraph{Single pass.}
After one pass, the fused accuracy is 73.8\% (Figure~\ref{fig:online}, left), 3.9\% below six offline passes and still above MISNet-S and SMMT, which train for many epochs. The network recruits more segments as it learns, from 0.19M synapses after 100 clips to 3.87M after the pass.

\paragraph{Adapting to new actors.}
Here, we assess whether the network can adapt to a new speaker from a few labeled clips. The network first learns online from 80\% of the clips of 60 actors. The other 31 actors are then presented one at a time. Half of each new actor's clips are used for learning and the other half for testing, and we measure the test accuracy on that actor after the network has learned from $k$ of its clips.
As a baseline, a copy of the network is frozen after the first 60 actors: its weights never change again, but it receives the same label-free normalization statistics of each new actor. Any difference between the two networks therefore comes from STDP learning, not from adapting the normalization.

Unseen actors start near 40\% and gain on average 19\% after 20 labeled clips and 32\% after all of them, about 41 clips (Figure~\ref{fig:online}, right). The frozen baseline stays flat near 40\%, so the gain comes from the STDP updates.
Re-testing the first 60 actors on their held-out clips afterwards costs 5.4\% of fused accuracy. 
The class layers lose only 0.2--3.4\%, because new actors mostly recruit idle segments; the rest of the loss comes from the 108-synapse decision layer re-tuning to recent actors.

\section{Ablation Study}
\label{sec:ablation}

Table~\ref{tab:ablation} (ablation study) removes one component at a time.

\paragraph{Recruit-on-error is effective.}
Without it, a class keeps the segments recruited only when it is entirely silent, and each segment averages many dissimilar samples. Accuracy falls by 21--23\% on both UrbanSound8K-AV layers and by 33\% on the CREMA-D faces. The exception is CREMA-D audio, where samples of an emotion vary so much across speakers that averaged templates generalise as well as stored ones.


\paragraph{Decision coding.}
Temporal coding in the decision layer ties about a fifth of the samples and loses 16\% on UrbanSound8K-AV and 4.6\% on CREMA-D. The rank-order code recovers the information that a floating-point weighted sum of scores exploits, with 4-bit weights: it is 1.7\% above the sum on UrbanSound8K-AV and 1.8\% above it on CREMA-D.

\begin{figure}[!t]
\centering
\includegraphics[width=\columnwidth]{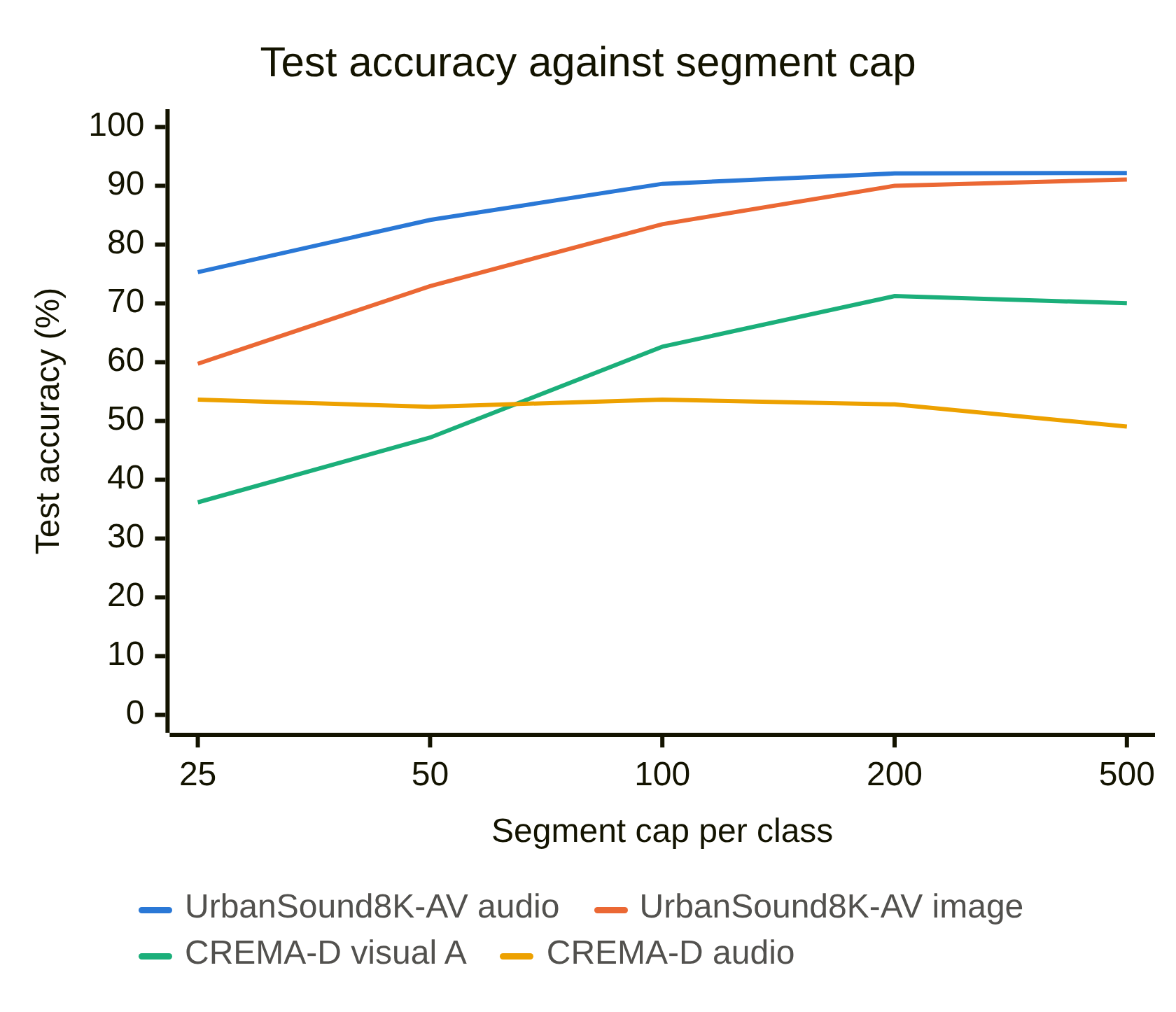}
\caption{Test accuracy of single class layers against segments per class, from 25 to 500. UrbanSound8K-AV layers stop recruiting at a cap of about 300, so larger caps add no segments. CREMA-D audio
peaks at 25--50 segments per class, and the CREMA-D visual layer at 200.}
\label{fig:caps}
\end{figure}

\paragraph{Segment cap.}
Figure~\ref{fig:caps} varies the segment count cap $K$ on each class layer. Accuracy rises steeply up to a few hundred segments per class and then flattens. On UrbanSound8K-AV, the curve ends because recruit-on-error stops recruiting. With a cap of 500, the audio and image layers recruit exactly as many segments as with 300, so the layer size is set by the data, not by the cap. Recruitment also distributes segments by need: among classes with the same number of training clips, the audio layer gives 79 segments to \emph{air conditioner} and 98 to \emph{engine idling}, both stationary sounds, but 215 to \emph{dog bark} and 190 to \emph{street music}, whose recordings vary far more. On CREMA-D, the face layers fill their caps and benefit from them up to about 200 segments per class, whereas the audio layer is as accurate with 25 segments per class (0.09M synapses) as with 200.


\section{Discussion}

\paragraph{Why RNL with input-normalized threshold is a good classifier.}
The capped ramp of Eq.~\eqref{eq:rnl} makes each synapse contribute at most its weight, so one unusually strong feature cannot dominate a segment. The input-normalized threshold of Eq.~\eqref{eq:thr} makes the race compare patterns rather than magnitudes, and WTA over segments implements a nearest-match decision in time, read out by the first spike. These operations need only comparators and small counters on 4-bit weights. Together with the small synapse count, this gives the 23--106$\times$ memory reduction and removes per-synapse multiplications.

\paragraph{Fusion in a spiking decision layer.}
The two modalities of UrbanSound8K-AV make largely independent errors, and combining class evidence is enough to gain 6.3\% (7.4 on CREMA-D); the decision layer does this with at most 400 synapses. The lesson from the decision coding ablation is that the code between layers matters more than the decision layer's size: ranks preserve the ordering of runner-up classes that a coarse temporal code destroys.

\begin{table}[t]
\centering\small
\setlength{\tabcolsep}{0.5pt}
\caption{Ablations (accuracy, \%). UB: UrbanSound8K-AV, CR: CREMA-D.}
\label{tab:ablation}
\begin{tabular}{lcccc}
\toprule
Class layer & UB aud. & UB vis. & CR aud. & CR vis. \\
\midrule
Full model & 92.44 & 92.52 & 52.42 & 71.24 \\
No recruit-on-error & 70.69 & 67.82 & 52.69 & 37.23 \\
\midrule
Decision layer & \multicolumn{2}{c}{UB fused} & \multicolumn{2}{c}{CR fused} \\
\midrule
Rank code (4-bit) & \multicolumn{2}{c}{98.82} & \multicolumn{2}{c}{78.63} \\
Temporal code of $z$-scores & \multicolumn{2}{c}{82.86} & \multicolumn{2}{c}{74.06} \\
Float weighted sum (non-spiking) & \multicolumn{2}{c}{97.14} & \multicolumn{2}{c}{76.88} \\
\bottomrule
\end{tabular}
\end{table}

\paragraph{Connections to neuroscience.}
Three ingredients of the model have direct biological counterparts. The proximal teaching input follows theories in which a separate dendritic compartment receives an instructive signal that gates plasticity of the synapses carrying sensory input \citep{guerguiev2017towards,sacramento2018dendritic,payeur2021burst}. Here, the compartment is the proximal input and the signal is a class label, delivered only during learning. The reward-signed update is a three-factor rule of the kind attributed to neuromodulators \citep{izhikevich2007solving,fremaux2016neuromodulated}. Recruit-on-error is a form of structural plasticity: cortical circuits form new spines and synapses during learning \citep{holtmaat2009structural}, and our neurons grow a new distal segment when the existing ones fail. Finally, the retina transmits information in the relative latencies of first spikes \citep{gollisch2008rapid}, and rank-order codes were proposed as a fast cortical code on that basis \citep{thorpe1998rank}. Our ablation shows the same code is what lets a small spiking layer combine two senses.


\section{Conclusion}

In this work, we extend the neuromorphic NeuTNN model and propose NeuTNN-AV to support multimodal audio-visual classification with RNL class layer(s) per modal stream and an RNL decision layer. All layers are trained by local reward-modulated STDP with error-driven recruitment of dendritic segments. With 4-bit synapses and no gradient-based back-propagation training, NeuTNN-AV exceeds MISNet-XL accuracy on CREMA-D and comes within 0.14\% of it on UrbanSound8K-AV with 13-23$\times$ fewer synapses. It needs 23--106$\times$ less weight memory and 2-3 orders of magnitude less synaptic energy. Because learning is local, the same network learns in a single pass and adapts to new actors on the fly. 

This work clearly demonstrates the capability and efficiency of the NeuroAI Temporal Neural Networks, with active dendrites and distal and proximal segments, for multi-modal audio-visual classification with potential for continual online learning. This work also demonstrates the benefit of multimodal fusion to achieve better classification than single-modality approaches. Our future research will focus on other combinations of sensory modalities and fully developing online continuous learning for compelling applications.
\bibliography{aaai2027}

\begin{thebibliography}{45}
\providecommand{\natexlab}[1]{#1}

\bibitem[{Bi and Poo(1998)}]{bi1998synaptic}
Bi, G.-q.; and Poo, M.-m. 1998.
\newblock Synaptic Modifications in Cultured Hippocampal Neurons: Dependence on Spike Timing, Synaptic Strength, and Postsynaptic Cell Type.
\newblock \emph{Journal of Neuroscience}, 18(24): 10464--10472.

\bibitem[{{BrainChip Holdings Ltd.}(2025)}]{brainchip}
{BrainChip Holdings Ltd.} 2025.
\newblock \emph{{AKD1000 Akida System-on-Chip Product Brief}}.
\newblock Version 2.3.

\bibitem[{Cao et~al.(2014)Cao, Cooper, Keutmann, Gur, Nenkova, and Verma}]{cao2014cremad}
Cao, H.; Cooper, D.~G.; Keutmann, M.~K.; Gur, R.~C.; Nenkova, A.; and Verma, R. 2014.
\newblock Crema-d: Crowd-sourced emotional multimodal actors dataset.
\newblock \emph{IEEE transactions on affective computing}, 5(4): 377--390.

\bibitem[{Carandini and Heeger(2012)}]{carandini2012normalization}
Carandini, M.; and Heeger, D.~J. 2012.
\newblock Normalization as a canonical neural computation.
\newblock \emph{Nature reviews neuroscience}, 13(1): 51--62.

\bibitem[{Carpenter and Grossberg(1986)}]{carpenter1986adaptive}
Carpenter, G.~A.; and Grossberg, S. 1986.
\newblock Adaptive resonance theory: Stable self-organization of neural recognition codes in response to arbitrary lists of input patterns.
\newblock In \emph{Proceedings of the Annual Meeting of the Cognitive Science Society}, volume~8.

\bibitem[{Chaudhari et~al.(2021)Chaudhari, Nair, Moura, and Shen}]{chaudhari2021unsupervised}
Chaudhari, S.; Nair, H.; Moura, J.~M.; and Shen, J.~P. 2021.
\newblock Unsupervised clustering of time series signals using neuromorphic energy-efficient temporal neural networks.
\newblock In \emph{ICASSP 2021-2021 IEEE International Conference on Acoustics, Speech and Signal Processing (ICASSP)}, 7873--7877. IEEE.

\bibitem[{Dalal and Triggs(2005)}]{dalal2005hog}
Dalal, N.; and Triggs, B. 2005.
\newblock Histograms of oriented gradients for human detection.
\newblock In \emph{2005 IEEE computer society conference on computer vision and pattern recognition (CVPR'05)}, volume~1, 886--893. Ieee.

\bibitem[{Davies et~al.(2018)}]{davies2018loihi}
Davies, M.; et~al. 2018.
\newblock Loihi: A Neuromorphic Manycore Processor with On-Chip Learning.
\newblock \emph{IEEE Micro}, 38(1): 82--99.

\bibitem[{Davis and Mermelstein(1980)}]{davis1980mfcc}
Davis, S.; and Mermelstein, P. 1980.
\newblock Comparison of parametric representations for monosyllabic word recognition in continuously spoken sentences.
\newblock \emph{IEEE transactions on acoustics, speech, and signal processing}, 28(4): 357--366.

\bibitem[{De~Cheveign{\'e} and Kawahara(2002)}]{decheveigne2002yin}
De~Cheveign{\'e}, A.; and Kawahara, H. 2002.
\newblock YIN, a fundamental frequency estimator for speech and music.
\newblock \emph{The Journal of the Acoustical Society of America}, 111(4): 1917--1930.

\bibitem[{Diehl and Cook(2015)}]{diehl2015unsupervised}
Diehl, P.~U.; and Cook, M. 2015.
\newblock Unsupervised learning of digit recognition using spike-timing-dependent plasticity.
\newblock \emph{Frontiers in computational neuroscience}, 9: 99.

\bibitem[{Fan et~al.(2023)Fan, Xu, Wang, Wang, and Guo}]{fan2023pmr}
Fan, Y.; Xu, W.; Wang, H.; Wang, J.; and Guo, S. 2023.
\newblock Pmr: Prototypical modal rebalance for multimodal learning.
\newblock In \emph{2023 IEEE/CVF Conference on Computer Vision and Pattern Recognition (CVPR)}, 20029--20038. IEEE.

\bibitem[{Fang et~al.(2021)Fang, Yu, Chen, Huang, Masquelier, and Tian}]{fang2021sew}
Fang, W.; Yu, Z.; Chen, Y.; Huang, T.; Masquelier, T.; and Tian, Y. 2021.
\newblock Deep residual learning in spiking neural networks.
\newblock \emph{Advances in neural information processing systems}, 34: 21056--21069.

\bibitem[{Fr{\'e}maux and Gerstner(2016)}]{fremaux2016neuromodulated}
Fr{\'e}maux, N.; and Gerstner, W. 2016.
\newblock Neuromodulated Spike-Timing-Dependent Plasticity, and Theory of Three-Factor Learning Rules.
\newblock \emph{Frontiers in Neural Circuits}, 9: 85.

\bibitem[{Gollisch and Meister(2008)}]{gollisch2008rapid}
Gollisch, T.; and Meister, M. 2008.
\newblock Rapid neural coding in the retina with relative spike latencies.
\newblock \emph{science}, 319(5866): 1108--1111.

\bibitem[{Guerguiev, Lillicrap, and Richards(2017)}]{guerguiev2017towards}
Guerguiev, J.; Lillicrap, T.~P.; and Richards, B.~A. 2017.
\newblock Towards Deep Learning with Segregated Dendrites.
\newblock \emph{eLife}, 6: e22901.

\bibitem[{Guo et~al.(2023)Guo, Gao, Qu, Zheng, Jiang, Lu, and Qiao}]{guo2024smmt}
Guo, L.; Gao, Z.; Qu, J.; Zheng, S.; Jiang, R.; Lu, Y.; and Qiao, H. 2023.
\newblock Transformer-based spiking neural networks for multimodal audiovisual classification.
\newblock \emph{IEEE Transactions on Cognitive and Developmental Systems}, 16(3): 1077--1086.

\bibitem[{Hawkins and Ahmad(2016)}]{hawkins2016neurons}
Hawkins, J.; and Ahmad, S. 2016.
\newblock Why neurons have thousands of synapses, a theory of sequence memory in neocortex.
\newblock \emph{Frontiers in neural circuits}, 10: 23.

\bibitem[{Holtmaat and Svoboda(2009)}]{holtmaat2009structural}
Holtmaat, A.; and Svoboda, K. 2009.
\newblock Experience-dependent structural synaptic plasticity in the mammalian brain.
\newblock \emph{Nature Reviews Neuroscience}, 10(9): 647--658.

\bibitem[{Horowitz(2014)}]{horowitz2014energy}
Horowitz, M. 2014.
\newblock 1.1 computing's energy problem (and what we can do about it).
\newblock In \emph{2014 IEEE international solid-state circuits conference digest of technical papers (ISSCC)}, 10--14. Ieee.

\bibitem[{Izhikevich(2007)}]{izhikevich2007solving}
Izhikevich, E.~M. 2007.
\newblock Solving the distal reward problem through linkage of STDP and dopamine signaling.
\newblock \emph{Cerebral cortex}, 17(10): 2443--2452.

\bibitem[{Jiang et~al.(2002)Jiang, Lu, Zhang, Tao, and Cai}]{jiang2002contrast}
Jiang, D.-N.; Lu, L.; Zhang, H.-J.; Tao, J.-H.; and Cai, L.-H. 2002.
\newblock Music type classification by spectral contrast feature.
\newblock In \emph{Proceedings. IEEE international conference on multimedia and expo}, volume~1, 113--116. IEEE.

\bibitem[{Kheradpisheh et~al.(2018)Kheradpisheh, Ganjtabesh, Thorpe, and Masquelier}]{kheradpisheh2018stdp}
Kheradpisheh, S.~R.; Ganjtabesh, M.; Thorpe, S.~J.; and Masquelier, T. 2018.
\newblock STDP-based spiking deep convolutional neural networks for object recognition.
\newblock \emph{Neural Networks}, 99: 56--67.

\bibitem[{Liu et~al.(2022)Liu, Xing, Feng, Tang, and Pan}]{liu2022emsnn}
Liu, Q.; Xing, D.; Feng, L.; Tang, H.; and Pan, G. 2022.
\newblock Event-based multimodal spiking neural network with attention mechanism.
\newblock In \emph{ICASSP 2022-2022 IEEE International Conference on Acoustics, Speech and Signal Processing (ICASSP)}, 8922--8926. IEEE.

\bibitem[{Liu et~al.(2025)Liu, Xia, Zhou, Li, and Guo}]{liu2025misnet}
Liu, X.; Xia, N.; Zhou, J.; Li, Z.; and Guo, D. 2025.
\newblock Towards energy-efficient audio-visual classification via multimodal interactive spiking neural network.
\newblock \emph{ACM Transactions on Multimedia Computing, Communications and Applications}, 21(5): 1--24.

\bibitem[{Maass(1997)}]{maass1997networks}
Maass, W. 1997.
\newblock Networks of Spiking Neurons: The Third Generation of Neural Network Models.
\newblock \emph{Neural Networks}, 10(9): 1659--1671.

\bibitem[{Masquelier and Thorpe(2007)}]{masquelier2007unsupervised}
Masquelier, T.; and Thorpe, S.~J. 2007.
\newblock Unsupervised Learning of Visual Features through Spike Timing Dependent Plasticity.
\newblock \emph{PLoS Computational Biology}, 3(2): e31.

\bibitem[{Merolla et~al.(2014)Merolla, Arthur, Alvarez-Icaza, Cassidy, Sawada, Akopyan, Jackson, Imam, Guo, Nakamura et~al.}]{merolla2014million}
Merolla, P.~A.; Arthur, J.~V.; Alvarez-Icaza, R.; Cassidy, A.~S.; Sawada, J.; Akopyan, F.; Jackson, B.~L.; Imam, N.; Guo, C.; Nakamura, Y.; et~al. 2014.
\newblock A million spiking-neuron integrated circuit with a scalable communication network and interface.
\newblock \emph{Science}, 345(6197): 668--673.

\bibitem[{Mozafari et~al.(2018)Mozafari, Kheradpisheh, Masquelier, Nowzari-Dalini, and Ganjtabesh}]{mozafari2018first}
Mozafari, M.; Kheradpisheh, S.~R.; Masquelier, T.; Nowzari-Dalini, A.; and Ganjtabesh, M. 2018.
\newblock First-spike-based visual categorization using reward-modulated STDP.
\newblock \emph{IEEE transactions on neural networks and learning systems}, 29(12): 6178--6190.

\bibitem[{Nair, Shen, and Smith(2021)}]{nair2021microarchitecture}
Nair, H.; Shen, J.~P.; and Smith, J.~E. 2021.
\newblock A Microarchitecture Implementation Framework for Online Learning with Temporal Neural Networks.
\newblock In \emph{IEEE Computer Society Annual Symposium on VLSI (ISVLSI)}.

\bibitem[{Nair et~al.(2022)Nair, Vellaisamy, Bhasuthkar, and Shen}]{nair2022tnn7}
Nair, H.; Vellaisamy, P.; Bhasuthkar, S.; and Shen, J.~P. 2022.
\newblock Tnn7: A custom macro suite for implementing highly optimized designs of neuromorphic tnns.
\newblock In \emph{2022 IEEE Computer Society Annual Symposium on VLSI (ISVLSI)}, 152--157. IEEE.

\bibitem[{Neftci, Mostafa, and Zenke(2019)}]{neftci2019surrogate}
Neftci, E.~O.; Mostafa, H.; and Zenke, F. 2019.
\newblock Surrogate gradient learning in spiking neural networks: Bringing the power of gradient-based optimization to spiking neural networks.
\newblock \emph{IEEE Signal Processing Magazine}, 36(6): 51--63.

\bibitem[{Payeur et~al.(2021)Payeur, Guerguiev, Zenke, Richards, and Naud}]{payeur2021burst}
Payeur, A.; Guerguiev, J.; Zenke, F.; Richards, B.~A.; and Naud, R. 2021.
\newblock Burst-Dependent Synaptic Plasticity Can Coordinate Learning in Hierarchical Circuits.
\newblock \emph{Nature Neuroscience}, 24: 1010--1019.

\bibitem[{Poirazi, Brannon, and Mel(2003)}]{poirazi2003pyramidal}
Poirazi, P.; Brannon, T.; and Mel, B.~W. 2003.
\newblock Pyramidal neuron as two-layer neural network.
\newblock \emph{Neuron}, 37(6): 989--999.

\bibitem[{Reilly, Cooper, and Elbaum(1982)}]{reilly1982neural}
Reilly, D.~L.; Cooper, L.~N.; and Elbaum, C. 1982.
\newblock A neural model for category learning.
\newblock \emph{Biological cybernetics}, 45(1): 35--41.

\bibitem[{Sacramento et~al.(2018)Sacramento, Ponte~Costa, Bengio, and Senn}]{sacramento2018dendritic}
Sacramento, J.; Ponte~Costa, R.; Bengio, Y.; and Senn, W. 2018.
\newblock Dendritic Cortical Microcircuits Approximate the Backpropagation Algorithm.
\newblock In \emph{Advances in Neural Information Processing Systems (NeurIPS)}.

\bibitem[{Salamon, Jacoby, and Bello(2014)}]{salamon2014urbansound}
Salamon, J.; Jacoby, C.; and Bello, J.~P. 2014.
\newblock A dataset and taxonomy for urban sound research.
\newblock In \emph{Proceedings of the 22nd ACM international conference on Multimedia}, 1041--1044.

\bibitem[{Smith(2018)}]{smith2018spacetime}
Smith, J.~E. 2018.
\newblock Space-Time Algebra: A Model for Neocortical Computation.
\newblock In \emph{ACM/IEEE International Symposium on Computer Architecture (ISCA)}, 289--300.

\bibitem[{Smith(2020)}]{smith2020temporal}
Smith, J.~E. 2020.
\newblock A Temporal Neural Network Architecture for Online Learning.
\newblock ArXiv:2011.13844.

\bibitem[{Thorpe and Gautrais(1998)}]{thorpe1998rank}
Thorpe, S.; and Gautrais, J. 1998.
\newblock Rank order coding.
\newblock In \emph{Computational Neuroscience: Trends in Research, 1998}, 113--118. Springer.

\bibitem[{Venkatachalam et~al.(2026)Venkatachalam, Vellaisamy, Nair, Huang, Na, Kang, Jacobson, and Shen}]{venkatachalam2026neuroai}
Venkatachalam, S.; Vellaisamy, P.; Nair, H.; Huang, W.-C.; Na, Y.; Kang, Y.; Jacobson, Q.; and Shen, J.~P. 2026.
\newblock NeuroAI Temporal Neural Networks (NeuTNNs): Microarchitecture and Design Framework for Specialized Neuromorphic Processing Units.
\newblock \emph{arXiv preprint arXiv:2602.01546}.

\bibitem[{Viola and Jones(2001)}]{viola2001rapid}
Viola, P.; and Jones, M. 2001.
\newblock Rapid object detection using a boosted cascade of simple features.
\newblock In \emph{Proceedings of the 2001 IEEE computer society conference on computer vision and pattern recognition. CVPR 2001}, volume~1, I--I. Ieee.

\bibitem[{Zador et~al.(2023)Zador, Escola, Richards, {\"O}lveczky, Bengio, Boahen, Botvinick, Chklovskii, Churchland, Clopath et~al.}]{zador2023neuroai}
Zador, A.; Escola, S.; Richards, B.; {\"O}lveczky, B.; Bengio, Y.; Boahen, K.; Botvinick, M.; Chklovskii, D.; Churchland, A.; Clopath, C.; et~al. 2023.
\newblock Catalyzing next-generation artificial intelligence through neuroai.
\newblock \emph{Nature communications}, 14(1): 1597.

\bibitem[{Zhang et~al.(2023)Zhang, Wu, Zhang, Hu, Fu, Zhou, and Peng}]{zhang2023qmf}
Zhang, Q.; Wu, H.; Zhang, C.; Hu, Q.; Fu, H.; Zhou, J.~T.; and Peng, X. 2023.
\newblock Provable dynamic fusion for low-quality multimodal data.
\newblock In \emph{International conference on machine learning}, 41753--41769. PMLR.

\bibitem[{Zhou et~al.(2022)Zhou, Zhu, He, Wang, Yan, Tian, and Yuan}]{zhou2022spikformer}
Zhou, Z.; Zhu, Y.; He, C.; Wang, Y.; Yan, S.; Tian, Y.; and Yuan, L. 2022.
\newblock Spikformer: When spiking neural network meets transformer.
\newblock \emph{arXiv preprint arXiv:2209.15425}.

\end{thebibliography}

\end{document}